\documentclass[letterpaper, 10 pt, conference]{ieeeconf}  

\IEEEoverridecommandlockouts                              

\IEEEoverridecommandlockouts

\usepackage{booktabs}
\usepackage{multirow}
\usepackage{graphicx}
\usepackage{microtype}
\usepackage{amsmath,amssymb,amsfonts}
\usepackage{xcolor}
\usepackage{tikz}
\usepackage{textcomp}
\usepackage{lipsum}
\usepackage{pifont}
\usepackage[hidelinks]{hyperref}
\newcommand{\cmark}{\ding{51}}%
\newcommand{\xmark}{\ding{55}}%

\newcommand{\bb}[1]{{\textbf{#1}}}

\graphicspath{figures}

\title{\LARGE \bf
 Sim2Real Bilevel Adaptation for Object Surface Classification \\using Vision-Based Tactile Sensors
}

\author{Gabriele M. Caddeo$^{1, 3, \dag}$, Andrea Maracani$^{1, 3, 4, \dag}$, Paolo Didier Alfano$^{2, 4, \dag}$,\\ Nicola A. Piga$^{1}$, Lorenzo Rosasco$^{2, 4}$ and Lorenzo Natale$^{1}$%
\thanks{$^{1}$Humanoid Sensing and Perception, Istituto Italiano di Tecnologia,
Genoa, Italy. {\tt\small name.surname@iit.it}}%
\thanks{$^{2}$IIT@MIT, Istituto Italiano di Tecnologia, Genoa, Italy. {\tt\small name.surname@iit.it}}%
\thanks{$^{3}$DIBRIS, Universit\`a di Genova, Via All'Opera Pia, 13, Genoa, Italy.}%
\thanks{$^{4}$MaLGa Center, DIBRIS, Universit\`a di Genova, Genoa, Italy.}%
\thanks{$\dag$Equal contribution.}%
}

\begin{document}

\maketitle
\thispagestyle{empty}
\pagestyle{empty}

\begin{abstract}

In this paper, we address the Sim2Real gap in the field of vision-based tactile sensors for classifying object surfaces. We train a Diffusion Model to bridge this gap using a relatively small dataset of real-world images randomly collected from unlabeled everyday objects via the DIGIT sensor. Subsequently, we employ a simulator to generate images by uniformly sampling the surface of objects from the YCB Model Set. These simulated images are then translated into the real domain using the Diffusion Model and automatically labeled to train a classifier. During this training, we further align features of the two domains using an adversarial procedure. Our evaluation is conducted on a dataset of tactile images obtained from a set of ten 3D printed YCB objects. The results reveal a total accuracy of 81.9\%, a significant improvement compared to the 34.7\% achieved by the classifier trained solely on simulated images. This demonstrates the effectiveness of our approach. We further validate our approach using the classifier on a 6D object pose estimation task from tactile data.

\end{abstract}

\section{INTRODUCTION}
Perception of object properties is a fundamental requirement to accomplish everyday tasks. Humans usually have a sense of the object surfaces through vision but sometimes they have to integrate or substitute this information with tactile information. Knowing the kind of surface they are dealing with while manipulating an object can be of primary importance to understand the pose of the object or to decide the next actions. 
\par
Among the available tactile sensors, vision-based tactile sensors\cite{digit, gelsight} represent the state of the art when it comes to help robots perceive object properties as they produce high resolution RGB tactile images of the surface in contact. Such images can be exploited with Deep Learning techniques, despite that they require a huge amount of data for training, which can be difficult to collect in the real world. Although simulators \cite{tacto, taxim} have been proposed to overcome this issue, they hardly reproduce the effect of mechanical properties, e.g., gel deformation, and light distribution with high fidelity.
\par
In this work, we aim to fill the gap between simulated and real images by translating the simulated images to the real domain using a Diffusion Model (DM) trained on few real \emph{unlabeled} images coming from a DIGIT sensor \cite{digit}. 
We propose a surface classifier which can distinguish among four classes: flat, curve, edge and corner. 
The classifier is trained with simulated images that are sampled on the surface of objects from the YCB Model Set \cite{calli2015ycb} and then translated with the DM. To label the images, we sample a point cloud on the object mesh and we employ an \emph{automatic} procedure to evaluate, for each point, the local curvature from which we extract the labels. Notably, the overall procedure to generate training data does not require manual annotations, as it uses a small set of unlabelled data from real contacts and a larger dataset that is acquired in simulation and automatically labelled.
\par
\begin{figure}
    \centering
    \includegraphics[scale=0.4]{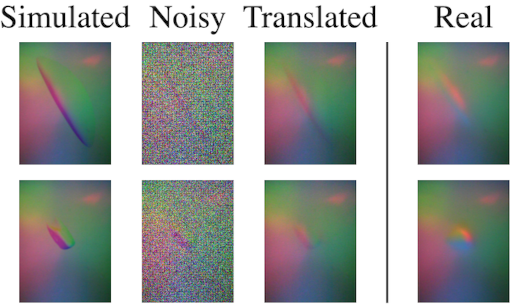}
    \caption{Our pipeline uses a Diffusion Model to translate simulated images towards the real domain so as to reduce the Sim2Real gap.\label{fig:diffusion_samples}}
    \vskip -1.5em
\end{figure}
We test the classifier on real tactile images acquired on ten YCB objects. We compare our model by training the classifier with simulated images or with images converted with an alternative DM-based model from the literature \cite{tactilediffusion} having more complex training requirements than ours. The experiments show that our method can achieve better accuracy in the classification task. Moreover, we employ our classifier within a pipeline for 6D object pose estimation \cite{caddeoicra2023} from multiple tactile sensors.

We summarize our contributions as follows:
\begin{itemize}
\item An image translation procedure to fill the Sim2Real gap using a Diffusion Model which easily generalizes to different vision-based tactile sensors; 
\item An object-agnostic surface type classifier trained using simulated images and few real images;
\item A method to automatically label the object surfaces given the object mesh, that we use to train the above classifier.
\end{itemize}
\begin{figure*}
        \vskip 0.2em
	\centering
	\includegraphics[scale=0.07]{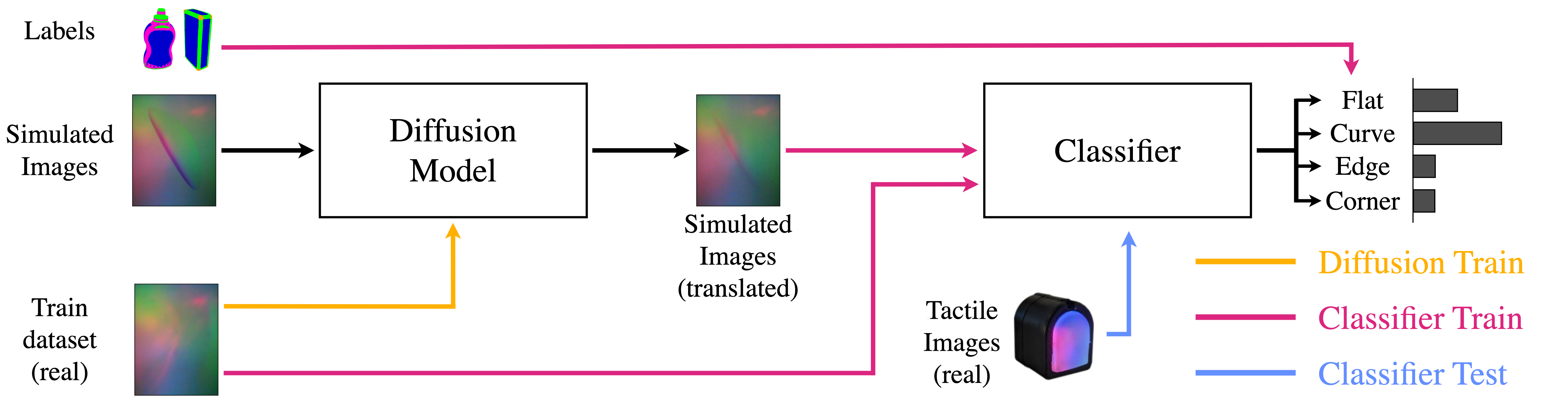}
	\caption{Overview of our pipeline for object surface classification.\label{fig:scheme}}
    \vskip -1,5em
\end{figure*}
\section{RELATED WORK}
This work draws from the literature on both Sim2Real translation for vision-based tactile sensors and perception of object properties using the same sensors.
\par
\textbf{Sim2Real}
Some work attempts to reduce the Sim2Real gap by carefully emulating data from real sensors. In \cite{gomes2021generation} and \cite{hogan2021seeing} the authors mimic the behaviour of the GelSight sensor using Phong's model. In \cite{tacto}, the authors present a simulator for the DIGIT sensor based on OpenGL using pyrender. A limitation of these methods is that they do not take into account the mechanical properties of the sensors. 
In \cite{taxim} the authors integrate the dependency on the mechanical properties of the gel by adding a calibration procedure based on the sensor output.
\par
Other approaches attempt to mitigate the domain shift between simulated and real images. In \cite{chen2022bidirectional}, the authors try to bridge the Sim2Real gap by employing a CycleGAN \cite{zhu2017unpaired} to simulate the complex light transmission observed in real sensors, a characteristic not captured by the simulator.  In contrast, our work utilizes simulated images, from TACTO \cite{tacto}, which undergo a transformation via a Diffusion Model. trained on real images, in order to mimic the real deformation of the gel and light transmission of the sensor. A similar approach is considered in \cite{tactilediffusion}, where, however, the Diffusion Model is trained with additional conditioning depth images that are obtained from a network~\cite{suresh2023midastouch} that had been previously trained. In our case, we dispense with the need for this network and rely solely on RGB images.
In the experimental section, we compare with a variant of our pipeline where the DM for image translation is substituted with that of \cite{tactilediffusion}.
\par
\textbf{Object perception with vision-based tactile sensors}
To the best of our knowledge, there is no work that directly classifies the surfaces of an object using vision-based tactile sensors. Nonetheless, we refer to other works which infer similar properties of the object.
\par
In \cite{xu2023visual} the authors integrate vision with touch to infer the shape of an object. In our work, no visual information is needed. In \cite{chen2023sliding} the authors train a network to retrieve the location and shape of the contacts in order to infer the shape of the object while the sensor slides along the object surface. 
In \cite{sodhi2022patchgraph}, the authors reconstruct the normals of the local surface of small objects thanks to a network trained with real and simulated images. On the contrary, we concentrate on objects whose size is not comparable to that of the sensor surface, thus producing more ambiguous tactile images and on general classes of surfaces. In \cite{suresh2023midastouch, caddeoicra2023} the authors identify the possible contact point on the object surface by comparing the input image with a database of images that are sampled on the object surface in advance, while our work does not need any database at inference time.

\section{METHOD}

Our approach leverages unlabeled real-world data and labeled simulated data in order to obtain good classification performance in real-world scenarios. To achieve this goal, we devise an automated method for acquiring and labeling synthetic data. We incorporate two levels of adaptation to mitigate domain shift and enhance performance. Specifically, we employ a probabilistic Diffusion Model to translate the simulated images. 
We further adapt the model features through an adversarial process using the Domain-Adversarial Training of Neural Networks (DANN) method \cite{ganin2016domain}. 
\par
We remark that we always remove the background signal of the DIGIT, i.e. its RGB output when not in contact with an object, from both real and simulated images. As DIGIT sensors can exhibit slight background variations owing to manufacturing differences, this ensures that the methodology remains agnostic to the specific background.

\par
This section offers a comprehensive overview of all the components illustrated in Fig. \ref{fig:scheme} within our pipeline.

\subsection{Acquisition and labeling of simulated data}\label{subsec:labeling}
We employ Poisson disk sampling \cite{4342600White} to extract uniformly distributed point clouds from object meshes augmented with surface normals. For each point we simulate the image produced by the DIGIT, using the simulator \cite{tacto}, while considering several rotations of the sensor around the normal direction and several penetration depths, so as to ensure variability in the collected data.

To automate the labeling process, we devised a simple yet effective algorithm to proficiently categorize each point within the point clouds as either \textit{flat}, \textit{curve}, \textit{edge}, or \textit{corner}.
Let suppose to have a point cloud of $M$ points: $P=\{p_i\}_{i=1}^{M}$, where $p_i \in \mathbb{R}^3$. For every point $p_j \in P$  the algorithm, which operates in four distinct steps, computes automatically the class of the point:
\begin{enumerate}
    \item The neighborhood $N_j$ of $p_j$ is computed: $N_j = \{p \in P : ||p_j - p||^2 \leq R\}$, where the radius $R$ is an hyperparameter of the algorithm. 
    \item We extract the ordered singular values of the $3 \times 3$ covariance matrix of the points in $N_{j}$, i.e., $\sigma_1^{(j)} \geq \sigma_2^{(j)} \geq \sigma_3^{(j)}$.
    \item We define the curvature level of the point $p_j$ as the ratio:
    \begin{equation} \label{eq:curvature}
        \text{Curv}(p_j) := \frac{\sigma_3^{(j)}}{\sigma_1^{(j)} + \sigma_2^{(j)} + \sigma_3^{(j)}} 
    \end{equation}
    This method provides an estimation of the local surface behavior. Intuitively, when the local curvature level is 0, the local area is entirely flat (as the smallest singular value $\sigma_3=0$) and the points would be distributed just on the plane defined by the first two eigenvectors corresponding to $\sigma_1$ and $\sigma_2$. Conversely, when the local curvature level is $1/3$ (the maximum value attainable by the curvature function), the surface displays pronounced curvature, as all three singular values are equal. By setting two thresholds ($0 < t_1 < t_2 < 1/3$) on the local curvature level, we are able to partition and classify all the points into three categories: \textit{flat}, \textit{curve}, and \textit{hard-curve}.
    \item Finally, points previously classified as \textit{hard-curve} are further separated into \textit{edges} and \textit{corners}, a task that cannot be achieved using curvature levels alone. In this step, we utilize K-means clustering on the \emph{normal} vectors of the neighborhood of \textit{hard-curve} points to classify them as either \textit{edges} or \textit{corners}. Intuitively, edges are defined by two flat surfaces represented by two normal vectors. Consequently, the normal vectors of the neighborhood of edge points should be easily clustered with $K=2$, and the advantage of using $K=3$ should be negligible. On the other hand, corner points are defined by three planes, and the normal vectors of the neighborhood cannot be easily clustered with $K=2$. Following this intuition, we compute the following quantity:
    \begin{equation}
        \Delta \text{loss}_{23} := loss(K=2) - loss(K=3),
    \end{equation}
    where $loss(K=q)$ is the final loss of K-Means when $K$ is set to $q$.
    A threshold on the value of $\Delta \text{loss}_{23}$ is set to partition \textit{edges} and \textit{corners}: low values of this metric identify edges, while high values identify corners. 
    
\end{enumerate}

We show the effectiveness of the overall approach for two YCB objects in Fig. \ref{fig:qual_curvature}.

\subsection{Image-level adaptation}\label{subsec:ima}
As simulated images and actual images acquired with a DIGIT sensor exhibit notable differences (see Fig. \ref{fig:diffusion_samples}), we propose an unsupervised image-to-image translation method to address the domain shift between these two domains. Our approach leverages the reverse process of a probabilistic diffusion model.
Diffusion models are latent variable models that involve two processes: the forward (diffusion) process and the reverse process. Let $x_0$ represent an image (in our work, a real image acquired with a DIGIT sensor). The forward process is a fixed Markov chain that gradually introduces Gaussian noise to $x_0$ over a predefined number of steps $T$, as determined by a variance schedule defined by the parameters $\{\beta_j\}_{j=1}^T$. Specifically, at any given time step $t \in {1, \ldots, T}$, the forward process adds Gaussian noise to the image according to the following transition:
\begin{figure}
	\centering
	\includegraphics[scale=0.75]{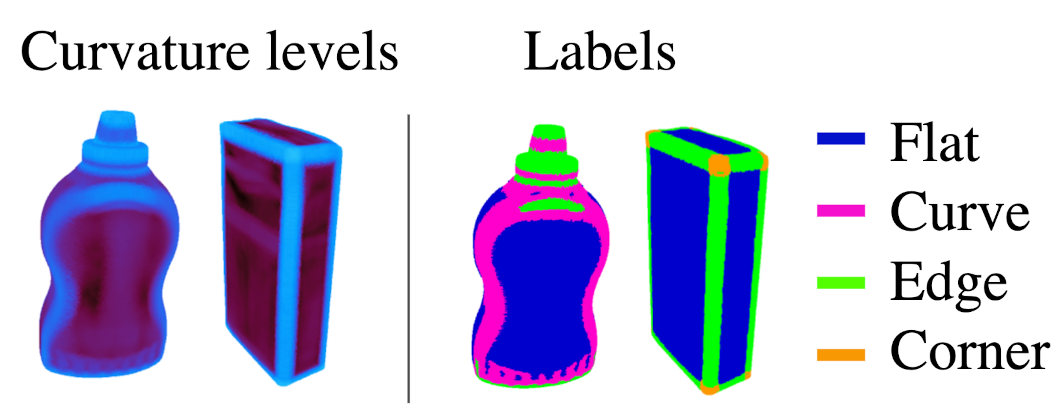}
	\caption{On the left: curvature levels (Eq. \ref{eq:curvature}) for the YCB objects ``mustard bottle'' and ``sugar box''. On the right: the resulting labels of surface types.\label{fig:qual_curvature}}
    \vskip -1.5em
\end{figure}
\begin{equation}
    q(x_t|x_{t-1}) := \mathcal{N}(x_t; \sqrt{1 - \beta_t} \cdot x_{t-1}, \beta_tI)
\end{equation}
where $\mathcal{N}$ is a Gaussian distribution and $I$ is the identity matrix.
The reverse process is also a Markov chain with Gaussian transitions that can be learned using a model, such as a neural network. This allows us to recover the image from the previous step ($x_{t-1}$) given the noisy image at step $t$ ($x_t$). By iterating this process, we can ultimately obtain the fully denoised image, $x_0$. To be more specific, the reverse transitions are as follows:
\begin{equation}
    p(x_{t-1}|x_{t}) := \mathcal{N}(x_{t-1}; \mu(x_t, t), \Sigma(x_t, t))
\end{equation}
Here, $\mu(\cdot, \cdot)$ and $\Sigma(\cdot, \cdot)$ represent functions that can be learned during the training process of the diffusion model. In our work, we utilized a U-Net architecture \cite{ronneberger2015u}, which is commonly employed in the literature.

Following the training, it becomes possible to sample an image with random Gaussian noise (bearing a strong resemblance to the images generated in the forward process at step $T$) and iteratively denoise it. This iterative process enables the generation of an image from the distribution of the training dataset. For more intuitions and for a more detailed description on probabilistic diffusion models the reader can refer to \cite{sohl2015deep} and \cite{ho2020denoising}.

To narrow the domain gap between simulated and real images, our approach encompasses the following steps:

\begin{enumerate}
    \item We train a U-Net model solely with unlabeled real images acquired with the DIGIT sensor to acquire knowledge of the reverse process.
    \item Once trained, we introduce a moderate level of noise to the simulated images, advancing in the forward process until step $T' < T$, with the aim of altering the \emph{style} of the images while preserving their semantic information.
    \item We employ the reverse process learned in step 1 to denoise the images, moving from step $T'$ to step $0$. As the diffusion model was trained using real images, the output comprises images that retain the semantic information of the simulated ones but exhibit a style more akin to real images as shown in Fig. \ref{fig:diffusion_samples}.
\end{enumerate}

Unlike previous works, such as \cite{tactilediffusion}, our diffusion method does not necessitate additional information during training or inference. It operates exclusively on raw images without any conditioning. The \textit{semantic conditioning} for image generation is accomplished, as mentioned earlier, by introducing a controlled level of noise into the simulated images, as it is similarly done in \cite{sdedit}. This preserves crucial information while preventing complete degradation.

\subsection{Feature-level adaptation}
Despite the significant reduction in domain shift achieved by the diffusion model, some residual differences between the domains still exist. To address this, we adopted a classical yet effective adversarial approach, known as Domain-Adversarial Training of Neural Networks (DANN), to facilitate the learning of domain-invariant representations.
\par
\begin{figure}
    \vskip 0.3em
    \centering
    \includegraphics[scale=0.085]{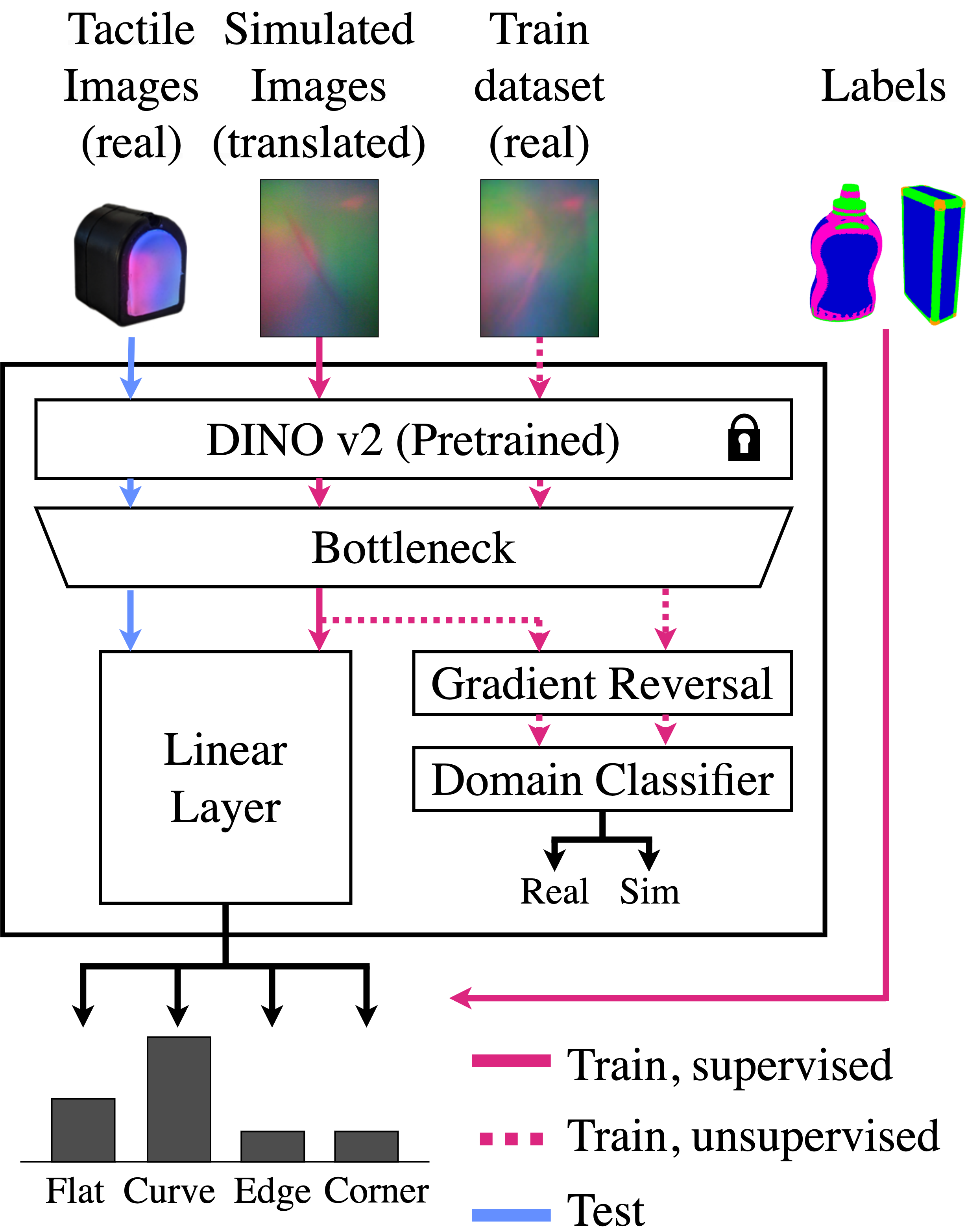}
    \caption{Diagram of the classification architecture.\label{fig:scheme_classifier}}
    \vskip -1.5em
\end{figure}

Specifically, we utilized a Vision Transformer (ViT) \cite{dosovitskiy2020image}, pretrained with DINO v2 method \cite{oquab2023dinov2}, as a feature extractor, keeping it fixed throughout the training process. We introduce and train a bottleneck layer (a linear layer, followed by Layer Normalization \cite{ba2016layer} and a GELU \cite{hendrycks2016gaussian} activation) to map ViT features into a domain-invariant space, along with a classifier that maps these bottleneck features to our target classes, as depicted in Fig. \ref{fig:scheme_classifier}. 
\par
During training, we employ a  discriminator to distinguish between real and simulated images, while the bottleneck layer is optimized to deceive the discriminator by making the features from both domains indistinguishable. To be more precise, we define $\phi(\cdot)$ to be our feature extractor (ViT) that maps images to features, $\beta(\cdot)$ as our bottleneck that reduces the dimension of features to $256$, $\gamma(\cdot)$ as a classifier that maps bottleneck features to our four classes and $\psi(\cdot)$ as a discriminator that maps bottleneck features to a domain label: $0$ for simulated images and $1$ for real images.
During training we sample a batch of real images (without labels) $x_{real}$ and a batch of simulated images (with labels) $(x_{sim}, y_{sim})$. We compute the bottleneck features ($\hat{z}$), the predictions of the network $\hat{y}_{sim}$ and the domain predictions of the discriminator $\hat{d}_{all}$ as follow: 

\begin{align}
    \hat{z}_{real} &= \phi(\beta(x_{real})) \\
    \hat{z}_{sim}  &= \phi(\beta(x_{sim})) \\
    \hat{y}_{sim}  &= \gamma(\hat{z}_{sim}) \\
    \hat{d}_{all}  &= \psi(\hat{z}_{all})
\end{align}
where $\hat{z}_{all}$ is the concatenation of $\hat{z}_{sim}$ and  $\hat{z}_{real}$ along the batch dimension of the two inputs. 

The bottleneck and the classifier are trained to minimize the following loss:
\begin{equation}
    L_{cls} = \text{CE}(y_{sim}, \hat{y}_{sim}) - \alpha \cdot \text{BCE}(d_{all}, \hat{d}_{all})
\end{equation}
where $d_{all}$ are all the domain labels (0s and 1s based on the domain of the inputs), CE is the cross-entropy loss, BCE is the binary cross-entropy loss and $\alpha$ is a hyperparameter that we set to $1.2$ in all of our experiments.

The discriminator, on the other hand, is trained to minimize: 
\begin{equation}
    L_{dis} = \alpha \cdot \text{BCE}(d_{all}, \hat{d}_{all})
\end{equation}
For this reason the training proceeds as an adversarial game between the bottleneck and the discriminator. In practice, we trained the whole network (with the exception of the feature extractor that is fixed) in an end-to-end fashion using a Gradient Reversal Layer (GRL) as proposed in DANN \cite{ganin2016domain} (refer to this work for more details).

\subsection{Training and testing datasets}

For our experiments, we collected three different datasets:

\begin{enumerate}

    \item The first dataset comprises $5,000$ real images acquired using a DIGIT sensor on the surfaces of randomly selected everyday objects, excluding all objects from the YCB Model Set \cite{calli2015ycb}. We refer to this \emph{unlabeled} dataset as $\textbf{Train}_{real}$.

    \item The second dataset consists of $50,000$ simulated images, $12,500$ per class,
    acquired from randomly selected points sampled on the meshes of ten YCB objects (\textit{cracker box, tomato soup can, tuna fish can, pudding box, gelatin box, banana, bleach cleanser, bowl, power drill, wood block}) and translated to the real domain as in Sec. \ref{subsec:ima}. We employ the algorithm described in Section \ref{subsec:labeling} to associate a ground truth label for each image, and we refer to this dataset as $\textbf{Train}_{sim}$.

    \item The third dataset includes $792$ real images acquired using our setup from ten 3D printed YCB objects (\textit{master chef can, sugar box, mustard bottle, tuna fish can, potted meat can, banana, pitcher base, bleach cleanser, bowl, power drill}) (see Fig. \ref{fig:setup}). For every tactile image we  manually label the type of surface depending on the actual type of contact. We use the latter as the ground-truth label for evaluation purposes. We refer to this dataset as $\textbf{Test}_{real}$.
    

\end{enumerate}

\begin{table}
\vskip 1.0em
\tiny
\scriptsize
\centering
\caption{Results of the classification experiments:\\accuracy for each object and averaged.\label{tab:classification}}
\begin{tabular}{| l | c | c | c | c | c | c|}
\hline
\textbf{Image transl.} & \multicolumn {2}{c|}{None} & \multicolumn {2}{c|}{Tactile Diff. \cite{tactilediffusion}} & \multicolumn {2}{c|}{Ours} \\
\hline
\textbf{Train w/ DANN} & \xmark & \cmark & \xmark & \cmark  & \xmark & \cmark \\
\hline

master chef can & 52.3\% & 41.2\%, & 58.7\% & 80.9\% & 66.6\% & \bb{92.0\%}\\
sugar box & 35.0\% & 35.0\% & 84.0\% & 71.0\% & 57.9\%& \bb{93.0\%}\\
mustard bottle & 41.0\% & 48.0\% & 34.0\% & 60.0\% & 63.0\%& \bb{71.0\%}\\
tuna fish can & 57.5\% &68.1\% & 59.0\%& \bb{80.3\%} & 74.2\%& 75.7\%\\
potted meat can & 35.0\% & 43.0\% & 54.0\%& 76.0\%& \bb{87.0\%}&84.0\%\\
banana & 2.4\% &29.2\% & 29.2\% & 7.3\%& 56.0\%& \bb{78.0\%}\\
pitcher base & 40.6\% & 39.0\% & 85.9\% & 90.6\% & 64.0\%& \bb{98.4\%}\\
bleach cleanser & 38.0\% & 43.0\% & 39.0\% & 64.0\% & 69.0\%& \bb{84.0\%}\\
bowl & 40.3\% & 43.5\% & 62.9\% & 88.7\% & 53.2\% & \bb{90.3\%}\\
power drill & 3.1\% & 33.3\% & 50.0\% & 38.54\% & 23.9\% & \bb{60.4\%}\\
\hline
\textbf{Mean} & 34.7\% & 42.4\% & 55.6\% & 66.6\% & 61.6\%& \bb{81.9\%}\\
\hline
\end{tabular}
\vskip -1.5em
\end{table}

To demonstrate the generalization capabilities of our approach, we intentionally include 5 overlapping objects (\textit{tuna fish can, banana, bleach cleanser, bowl, power drill}) in both the $\textbf{Train}_{sim}$ and $\textbf{Test}_{real}$ datasets. As a result, our algorithm performs well on objects seen during training and also on new objects.

We used the unlabeled $\textbf{Train}_{real}$ dataset for training the Diffusion Model, while the classifier is trained (with DANN) using both $\textbf{Train}_{sim}$ and the unlabeled $\textbf{Train}_{real}$. The testing phase was exclusively conducted on the $\textbf{Test}_{real}$ dataset.

\section{EXPERIMENTAL RESULTS} \label{sec:experiments_results}
We evaluate the performance of the classifier by considering the accuracy for each object and the F1-Score when analysing the results class-wise. 
In the results we also consider a variation of the pipeline where the proposed Diffusion Model is substituted with the state-of-the-art alternative \cite{tactilediffusion}. Moreover, we perform several ablation studies in order to investigate the role of the Diffusion Model and of the DANN procedure.
\par
Beyond the classification task, we apply our method within a pipeline for 6D object pose estimation from multiple tactile sensors \cite{caddeoicra2023}, to show its effectiveness on a practical task.
\par
The results of further experiments are provided in the supplementary video.
\par
\begin{figure}
    \centering
    \includegraphics[scale=0.9]{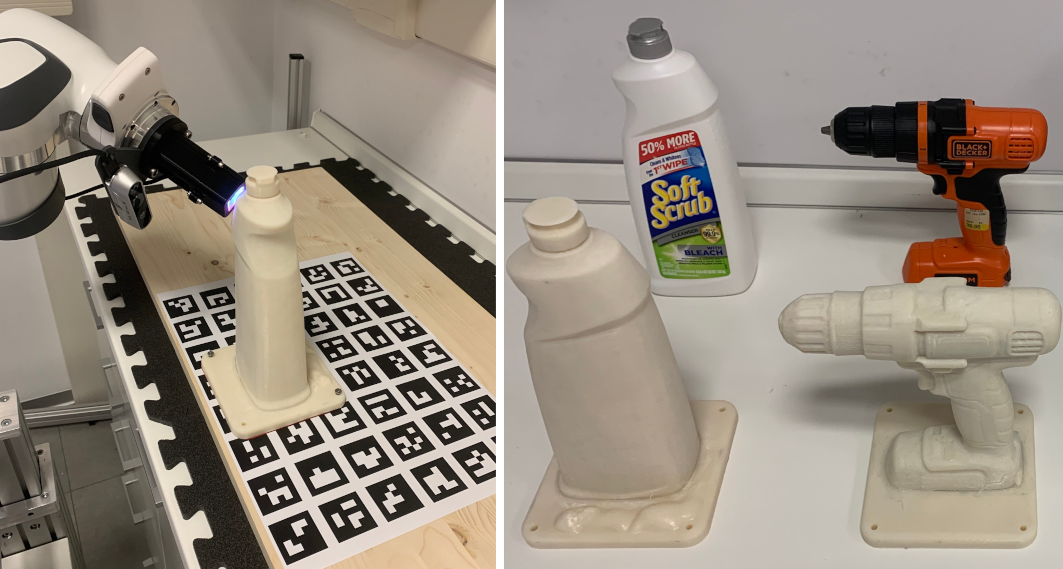}
    \caption{On the left: the testing setup with the DIGIT sensor touching the ``bleach cleanser'' object. On the right: example of YCB objects and their 3D printed counterparts.\label{fig:setup}}
    \vskip -1.5em
\end{figure}

In order to collect the $\textbf{Test}_{real}$  dataset, we use a DIGIT sensor mounted on a 7-DoF Franka Emika Panda robot to touch ten 3D printed YCB objects in several configurations. We show the setup and examples of the printed objects in Fig. \ref{fig:setup}.

\par

\par
In order to evaluate the outcome of the 6D object pose estimation experiments, we also collect the pose of the object in the robot root frame. To do so, we employ a fiducial system based on a RealSense camera and ArUco markers.
To make sure that the object does not move while touching it we fixed it to a table as shown in Fig. \ref{fig:setup}. We remark that the necessity to fix the objects is the only reason for printing them in 3D.
\par
Code and data will be made publicly available online\footnote{\href{https://github.com/hsp-iit/sim2real-surface-classification}{https://github.com/hsp-iit/sim2real-surface-classification}}.

\subsection{Experiments on surface classification}
Table \ref{tab:classification} presents the results in terms of \emph{accuracy} for each considered object. In Table \ref{tab:classification_surfaces}, we instead detail the results for each class using the F1-Score defined as follows:
\begin{equation*}
    \text{F1-Score} = \frac{2 \cdot \mathrm{precision} \cdot \mathrm{recall}}{\mathrm{precision} + \mathrm{recall}},
\end{equation*}
where \emph{precision} and \emph{recall} are defined as usual.
\begin{table}
\vskip 1.0em
\scriptsize
\centering
\caption{Results of the classification experiments:\\F1 score for each surface type.\label{tab:classification_surfaces}}
\begin{tabular}{| l | c | c | c | c | c | c|}
\hline
\textbf{Image transl.} & \multicolumn {2}{c|}{None} & \multicolumn {2}{c|}{Tactile Diff. \cite{tactilediffusion}} & \multicolumn {2}{c|}{Ours} \\
\hline
\textbf{Train w/ DANN} & \xmark & \cmark & \xmark & \cmark & \xmark & \cmark \\
\hline
flat & 46.5\% & 28.6\% & 61.5\% & 81.3\% & 86.4\% & \bb{91.1\%}\\
curve & 3.6\% & 31.1\% & 18.6\% & 30.3\% & 45.2\% & \bb{73.5\%}\\
edge & 28.\% & 56.2\% & 78.1\% & 74.4\% & 58.1\% & \bb{83.2\%}\\
corner & 0.0\% & 0.0\% & 46.6\% & 26.3\% & 21.1\% & \bb{68.3\%}\\
\hline
\end{tabular}
\end{table}

\par
In order to investigate how the image translation and the DANN procedure contribute to the performance we repeated the experiments using different training configurations of the pipeline. As regards the image translation we trained the classifier using:
\begin{itemize}
    \item simulated images, indicated as \textbf{``None''};
    \item translated images from the diffusion model proposed in \cite{tactilediffusion}, indicated as \textbf{``Tactile Diffusion''};
    \item translated images from the diffusion model proposed in this work, indicated as \textbf{``Ours''}.
\end{itemize}

For each configuration above, we repeated the experiments with the DANN procedure enabled or disabled.
\par
The results in Table \ref{tab:classification} shows that the best performance, except few cases, is achieved when using the proposed diffusion model as the translation mechanism combined with the DANN procedure.
\par
Using DANN, the proposed diffusion model increases the accuracy by $\approx 40$ points with respect to the ``None'' case and by $\approx 15$ points with respect to the method \cite{tactilediffusion}. 
\par
As regards the DANN procedure, on average it helps increase the accuracy in all the configurations. The best improvement, of $\approx 20$ points, is obtained in conjunction with the proposed diffusion model.
\par
Table \ref{tab:classification_surfaces} shows that the proposed pipeline achieves the best performance on all surface types. Remarkably, both the DM and DANN are essential to improve performance on all classes, and specifically to gain acceptable performance on the class corners.
\par
\begin{figure}
    \vskip 0.5em
    \centering
    \includegraphics[scale=0.9]{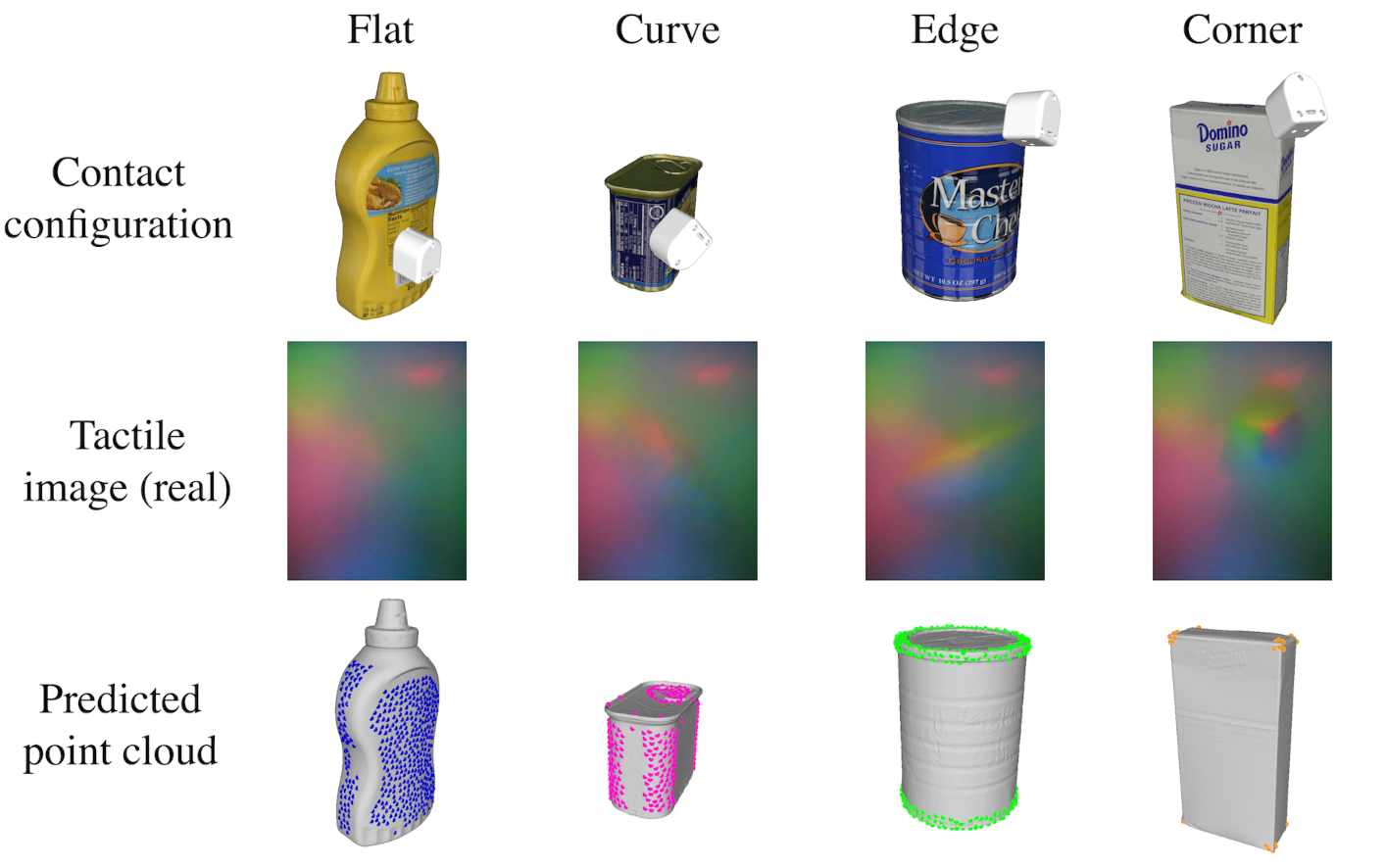}
    \caption{Visualization of the predicted surface type in terms of the object-sampled points having the same label as the predicted one.\label{fig:qual_contact_classification}}
    \vskip -1.5em
\end{figure}

In Fig. \ref{fig:qual_contact_classification} we present some qualitative results. Specifically, for each tactile image, we show the name of the predicted class and all the points sampled on the object mesh, as per Sec. \ref{subsec:labeling}, having the same label as the predicted one. These results indicate that a combination of the proposed classifier and the automatic labeling procedure of Sec. \ref{subsec:labeling} is useful to provide hypotheses on the contact location of the sensor on the object surface.

\subsection{Experiments on 6D object pose estimation}
For these experiments we make use of the algorithm presented in \cite{caddeoicra2023} that estimates the 6D pose of an object in contact with N tactile sensors given input tactile images and the pose of the sensors from the robot proprioception.
\par
Specifically, \cite{caddeoicra2023} extracts several hypotheses on the 3D location of each sensor on the object surface given the input tactile image. Then, it uses gradient-based optimization to fuse the hypotheses from all sensors resulting in a set of candidate 6D poses of the object.
\par
In this paper, we substitute the hypothesis extraction part, based on a convolutional autoencoder and a set of per-object databases of latent features, with the proposed classifier.

\par
We choose to run the experiments using $N = 3$ sensors. As in \cite{caddeoicra2023}, we considered four different configurations of sensors for every object under study.
\par
We compare the performance against a purely geometric baseline, as done in \cite{caddeoicra2023}. The baseline skips the hypothesis extraction part and executes the optimization assuming that the N sensors might touch the object everywhere.

\par
To evaluate the performance we compare the output pose with the ground truth pose in terms of the positional error and a variant of the ADI-AUC metric \cite{Xiang-RSS-18}, introduced in \cite{caddeoicra2023}, 
which reports on the rotational error. Remarkably, we report quantitative results of experiments run with real tactile images while \cite{caddeoicra2023} restricts the quantitative analysis to simulated data and only provides qualitative results on real images.

\par
Table \ref{tab:pose} shows that using the tactile feedback, in the form of the proposed classifier, helps halving the positional error and increasing the rotational metric by more than ten points, on average.
\par

\begin{figure}
    \vskip 0.5em
    \centering
    \includegraphics[scale=0.4]{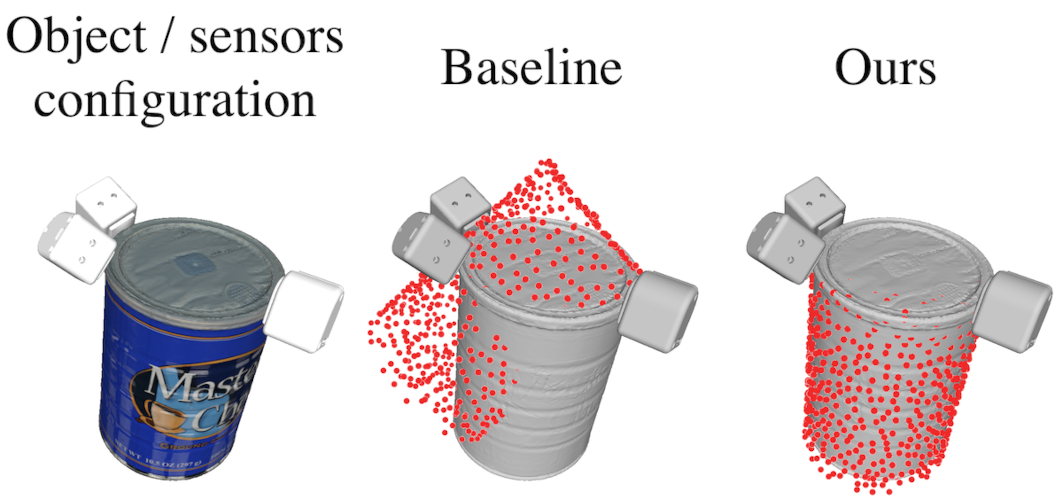}
\caption{Visualization of the outcome of the 6D object pose estimation experiment using real tactile images.\label{fig:qual_pose_estimation}}
\end{figure}

In Fig. \ref{fig:qual_pose_estimation} we show a sample outcome of the experiment and compare it against the baseline. The ground-truth pose is shown as the greyed-out mesh while the estimated pose as the red point cloud. As can be seen, although the pose estimated by the baseline is compatible with the \emph{position} of the contacts, two out of three sensors are in contact with the wrong part of the object. Instead, the pose estimated by our method is compatible with all contact positions and types. This further demonstrates the advantages of using the tactile feedback, in the form of the proposed classifier, for this task.

\begin{table}
\tiny
\scriptsize
\centering
\caption{Results of the object pose estimation experiments: averaged positional error and ADI-AUC with threshold set to $2\,\mathrm{cm}$.\label{tab:pose}}
\begin{tabular}{| l | c | c | c |c|}
\hline
\textbf{Metric} & \multicolumn {2}{c|}{Positional error (cm) $\downarrow$} & \multicolumn {2}{c|}{ADI-AUC$_{2 \mathrm{cm}}$ (\%) $\uparrow$}\\
\hline
\textbf{Method} & Baseline & Ours & Baseline & Ours \\
\hline

master chef can & 2.97 & \bb{0.69} & 93.54 & \bb{97.23}\\
sugar box & 3.61 & \bb{3.12} & \bb{93.75} & 72.61 \\
mustard bottle & 2.97 & \bb{1.83} & 69.83 & \bb{96.61} \\
tuna fish can & 2.18 &\bb{1.07} & \bb{96.08}& 95.97\\
potted meat can & 1.81 & \bb{1.26} & 94.71 & \bb{96.63}\\
banana & 6.72 &\bb{2.39} & 48.49 & \bb{96.93}\\
pitcher base & 6.71 & \bb{2.73} & 25.0 & \bb{47.19}\\
bleach cleanser & 4.97 & \bb{2.96} & 70.34 & \bb{94.28}\\
bowl & \bb{1.74} & 2.08 & \bb{97.29}& 96.27 \\
power drill & 8.36 & \bb{4.09} & 67.77 & \bb{73.76} \\

\hline
\textbf{Mean} & 4.21 & \bb{2.22} & 75.68 & \bb{86.75}\\
\hline
\end{tabular}
\vskip -1.5em
\end{table}

\section{Limitations}
The rigidity of the elastomer of the DIGIT sensor requires a more than moderate force, when interacting with objects, to highlight surface differences. Although this fact depends on the sensor itself, it might impact the effectiveness of our method if the contact forces are too weak.
\par
We acknowledge that our method makes use of a Diffusion Model whose training and image-translation times are non-negligible. Nonetheless, one of the advantages of the proposed method is that it is trained on images without background, thus it can be used on different devices without a re-train.

\section{Conclusion}
In this work, we tackled the Sim2Real gap in the context of vision-based tactile sensors to classify local surfaces of objects. Our method combines a bilevel adaptation, at the image and feature level, with an automatic labeling procedure, allowing us to train the classifier using a supervised approach while avoiding manual annotation. The extensive experiments on real data and the comparison against a state-of-the-art method validate the robustness and the effectiveness of our approach that we also applied successfully in the context of 6D object pose estimation from tactile data.
\par
As future work, we plan to exploit our classification approach for several other robotic tasks and to study new adaptation mechanisms to further increase the classification accuracy.

\section*{ACKNOWLEDGMENT}
We acknowledge financial support from the PNRR MUR project PE0000013-FAIR.



\addtolength{\textheight}{-8cm}   





\bibliography{surfacebib}

\end{document}